# CLASSIFICATION OF MOTORCYCLES USING EXTRACTED IMAGES OF TRAFFIC MONITORING VIDEOS

**Adriano Belletti Felicio**
**André Luiz Cunha**
Department of Transportation Engineering
School of Engineering of São Carlos
University of Sao Paulo

**ABSTRACT**
Due to the great growth of motorcycles in the urban fleet and the growth of the study on its behavior and of how this vehicle affects the flow of traffic becomes necessary the development of tools and techniques different from the conventional ones to identify its presence in the traffic flow and be able to extract your information. The article in question attempts to contribute to the study on this type of vehicle by generating a motorcycle image bank and developing and calibrating a motorcycle classifier by combining the LBP techniques to create the characteristic vectors and the classification technique LinearSVC to perform the predictions. In this way the classifier of vehicles of the type motorcycle developed in this research can classify the images of vehicles extracted of videos of monitoring between two classes motorcycles and non-motorcycles with a precision and an accuracy superior to 0,9.

## 1. INTRODUCTION

Due to the great growth of motorcycle participation in the urban fleet of several cities around the world, research has been done on the behavior of motorcyclists and how their presence interferes with traffic.

Conventional vehicle detection and information retrieval tools, such as floor sensors or monitoring cameras (Klein, 2001 ; Klein, Mills, & Gibson, 2006), have proved to be inefficient for motorcycles, mainly due to their physical and the peculiar way of moving between the other vehicles in the traffic chain (Yuan, Lu, & Sarraf, 1994).

KOV; YAI, (2009) developed a research to investigate the performance of urban traffic in cities where the fleet is predominantly motorcycles and the degree of their involvement in traffic accidents. BABU; VORTISCH; MATHEW, (2015) analyzed the movement patterns of motorcycles in mixed traffic using simulators to qualitatively reproduce the naturalistic behavior of motorcycles. NGUYEN; HANAOKA; KAWASAKI, (2014) on the other hand, observed traffic congestion in motorcyclists using safety space concepts, car-following movements and the use of virtual lanes. And in MUNIGETY; VICRAMAN; MATHEW, (2014) research developed a semiautomatic tool capable of extracting traffic data such as detailing the trajectory of several vehicles simultaneously using the traffic monitoring images. All these searches were done from video traffic monitoring images. Therefore, worldwide research has emphasized the importance of using computational vision concepts to investigate the behavior of motorcycles.

One of the techniques for extracting information from images is the creation of descriptors in a specific way, trying to identify characteristics of the objects of interest. The LBP (Local Binary Pattern) is a texture-based descriptor that obtains its data from a full scan across all pixels of the image by analyzing the variation of brightness between the reference pixel and its neighboring pixels. Usually this descriptor can use as a criterion of similarity the Euclidean distance or the Manhatan distance to recognize similar regions in different images (Guoying Zhao, Ahonen, Matas, & Pietikainen, 2012 ; Pietikäinen, Hadid, Zhao, & Ahonen, 2011 ; Takala, Ahonen, & Pietikäinen, 2005).

Meanwhile, the descriptor SURF (Speeded Up Robust Features) allows the detection and extraction of local characteristics, presenting reasonable efficiency even to small changes of perspective, rotation, scale, changes of lighting and noise in the images. The main factor that makes its performance better is its approximative model of scale space based on the integral image, allowing it to detect points by combing the image in its original size without the need for any pre-processing (Bay, Tuytelaars, & Van Gool, 2006 ; D. Kim & Dahyot, 2008). In a simplified way, the SURF algorithm can be separated into two parts: detect the points of interest and formulate a descriptor.

A classifier has the function of differentiating information into distinct groups or classes. The SVM classifier however is a technique that uses a set of input data called training, in this case images labeled with and without the object of interest. In the training set, significant information is extracted from the created descriptor, forming a characteristic vector and the image label. Thus, it is possible for the technique to perform the prediction / labeling on previously unseen images (Cao, Jiang, Cheng, & Wang, 2016 ; Fu, Ma, Liu, & Lu, 2016 ; D. Kim & Dahyot, 2008 ; Xiao, Gao, Kong, & Liu, 2014).

A tool to evaluate the prediction of a learning algorithm is the ROC (Receiver Operating Characteristic) curve analysis. This method considers the positive true rate (TVP), axis of the abscissa, and the false positive rate (TFP), axis of the ordinates, to represent the classification / prediction model by a point in the ROC space. Once you have represented the points in the ROC space of a model in question you can analyze the AUC (Area Under the ROC Curve) metric of this model. The AUC metric has gained prominence as a measure of data mining and machine learning models for evaluating the quality of the general prediction of the model in question (Fawcett, 2006 ; Prati, Batista, & Monard, 2008).

Therefore, this research aims to: (1) obtain a motorcycle image bank in traffic and (2) calibrate an SVM classifier for the detection of motorcycle type vehicles.

## 2. Proposed Method

Figure 1 details the steps of the proposed method.

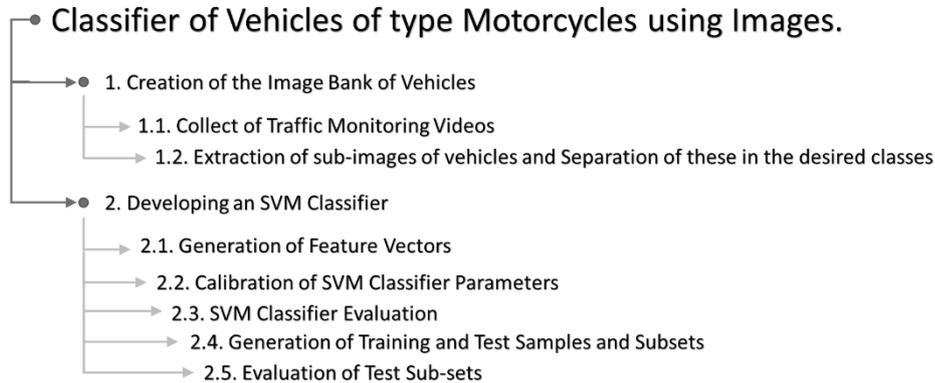

Figure 1: Scheme of the steps of the proposed method.

## 2.1. Creation of the Image Bank of Vehicles

The images present in the online repositories, such as *Imagenet*[1], *Pixabay*[2], *Compfight*[3], do not have many examples of vehicles in the traffic chain, most of them bring exhibition illustrations and vehicle stunts, highlighting the need to update these repositories. Thus, it was decided to create an own bank of images of vehicles acquired through videos of traffic monitoring collected by the authors.

*2.1.1. Collect of Traffic Monitoring Videos*

Data collection was performed on April 18, 2017 at Padre Francisco Salles Culturato Avenue in the West-East direction, in the city of Araraquara - Brazil. This route was chosen because it is considered one of the main interconnections of the city to the highway, besides presenting a large portion of motorcycles in its daily fleet. Two video cameras were used positioned on an existing walkway, one positioned in favor of the traffic flow, while the other recorded the opposite direction to the flow of traffic vehicles.

The filming period occurred between 10:50 a.m. and 2:10 p.m., period of intense flow of the avenue. In total, 15 videos of approximately 12 minutes each and 1 file of approximately 7 minutes were collected for each camera, all recorded in Full HD (1920 x 1080) at 30 fps.

*2.1.2. Extraction of sub-images of vehicles and Separation of these in the desired classes.*

From the videos collected, images of different types or categories of vehicles were extracted manually. It is worth noting a peculiarity observed, in the case of videos obtained in the direction in favor of the traffic flow, the vehicles are initially registered in large dimensions and as they follow their path there is a reduction of their size. In the case of videos obtained in the opposite direction of the traffic flow, the vehicles are initially sighted with reduced dimensions and as they follow their route, approaching the position where the camera is, there is an enlargement of its dimensions. Therefore, we chose to use in this research only the videos in favor of the flow.

To develop an image classifier it is necessary to standardize the size of the images (Corinna & Vapnik, 1995 ; Fu et al., 2016 ; Kotsiantis, Zaharakis, & Pintelas, 2006 ; R. Silva et al., 2013).

---

[1] Imagenet available on the site <https://image-net.org>
[2] Pixabay available on the site <https://pixabay.com>
[3] Compfight available on the site <https://compfight.com>

The standard size of 210x120 pixels was adopted, as they fitted the motorcycles completely in the initial moments of the videos in favor of the traffic flow.

To better define the vehicle framing, a standard sub-sample (210x120) was used throughout the frame, resulting in a 3x8 mesh. Figure 2 exemplifies the mesh used to extract the sub-images or vehicle cells in the videos.

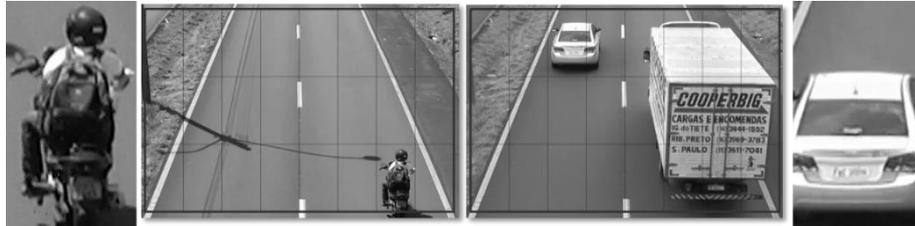

Figure 2: Representation of the projected mesh in the monitoring videos that registered the direction in favor of the current of traffic and examples of the extracted sub-imaging. 3x8 mesh producing 24 cells of 210 x 120 pixels each.

During manual identification, an observer recorded the presence and absence of vehicles in the videos. Only when the vehicles were as framed as possible within the cell, in the case of motorcycles, or within more than one cell, in the case of vehicles that were not motorcycles, the extraction of the sub - image or vehicle cell videos started.

Having the sub-images or vehicle cells extracted from the monitoring videos collected, it was possible to perform a manual classification of the images in the two interest classes: (i) motorcycles, containing 721 images and (ii) non-motorcycles, containing 13,393 images available of any other type of vehicle, part of vehicle or without vehicle.

Through this process, the classified vehicle image bank was created for the purpose of being used in the calibration of the classifier of motorcycle type vehicles.

**2.2. Developing an SVM Classifier**

The Support Vector Machine (SVM) classifier is a supervised mode Machine Learning technique. It has presented good performance in classification of images in general, of vehicles and also classification of motorcycles even when compared to other techniques (Cao et al., 2016 ; Fu et al., 2016 ; He, Du, Sun, & Wang, 2015 ; R. R. V. e Silva, Aires, & Veras, 2018 ; Xiao et al., 2014). The OpenCV and Scikit-learn libraries provide functions for this classifier

*2.2.1. Generation of Feature Vectors*

The input of any model of the type SVM occurs from vectors of characteristics used to the objects to be classified. Consequently, the input data to be adopted should always present the same amount of information for any element, unpublished or not, presented to it (Corinna & Vapnik, 1995 ; Pontil & Verri, 1998), stipulating that the characteristic vector a well-defined standard to facilitate and standardize the classification performed. The descriptors LBP (Local Binary Pattern) and SURF (Speeded Up Robust Features) were verified in the specialized literature.

In a simplified way, the SURF algorithm can be divided in two steps: (i) to detect the points of interest, with high variation of the direction intensity and (ii) to formulate a descriptor, from the extraction of local characteristics, presenting reasonable efficiency to small changes in perspective, rotation, scale, changes in lighting and noise in the images (Bay et al., 2006 ; Beyeler, 2015 ; D. Kim & Dahyot, 2008). However, the application of this technique in each image produces a set of points of interest, each point of this set containing a distinct amount of information. The technique allows to choose a set of 64 or 128 information per point of interest detected. However, to access to this information has been a very laborious task, making difficult to standardize the type of result provided through this approach, the set of information provided by this technique does not present a well-defined standard, rendering it inadequate as an input characteristics vector in the SVM model. Figure 3 shows the implementation of the SURF function in Python language available in the OpenCV library.

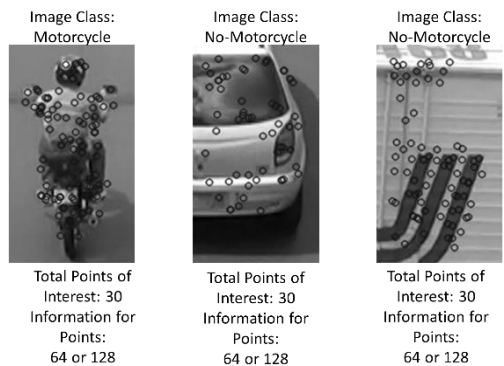

Figure 3 : The circles represent the regions of the images considered interesting (corners) by the SURF technique. In the image to the left we have a motorcycle, in the center a car and the right a truck.

In the case of LBP, the algorithm is based on texture, obtaining its data from a complete scan by all the pixels analyzing the variation of the brightness between the reference pixel and its neighboring pixels. This descriptor allows rotation invariance by counting the positive and negative variation between the analyzed pixels, i.e., the values of the pixels are thresholded by the value of the central pixel producing a threshold value, therefore, if the value of the surrounding pixel in the gray scale were greater than the value of the central pixel in the same gray scale, the result would be 1, otherwise the result would be 0. In this way the LBP code of the central pixel will be the sum of the product of the real image and the matrix of weights (Figure 4). The application of this technique in each image results in the texture of the image, with the same dimensions of the original image, which can be converted into a uniform and standard size histogram, as illustrated by Figure 4.

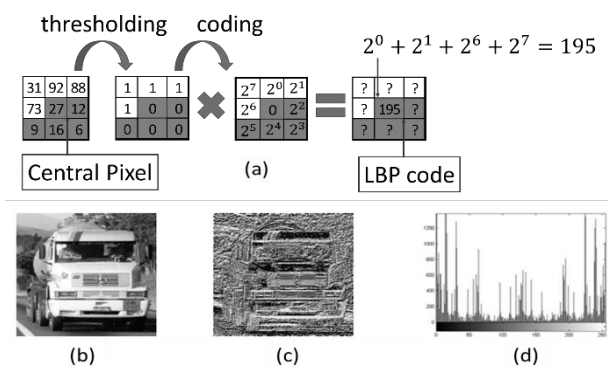

Figure 4: Example of the application of the LBP descriptor. (a) The procedure used by the descriptor. (b) Input image. (c) Result obtained. (d) Histogram produced by the descriptor. Adapted from CUNHA (2013).

Thus, the already normalized grayscale histogram provided by the technique to present a well-defined pattern can be used as a characteristic vector to be used as input data for LBP SVM type model. Figure 5 shows the implementation of the LBP function in Python language available in the scikit-image library with the respective parameters adopted to produce normalized grayscale histogram.

```
skimage.feature.local_binary_pattern(
                    image, "Graylevel image."
                    P = 3, "Number of circularly symmetric neighbour set points."
                    R = 8 * P, "Radius of circle (spatial resolution of the operator)"
                    method = "uniform" {"default", "ror", "uniform", "var"}, "Method to determine the pattern."
                    )
```

Figure 4: Parameters adopted for the LBP function.

### 2.2.2. . Calibration of SVM Classifier Parameters

The LinearSVC function is available in the scikit-learn library and 12 parameters must be configured (Figure 6). For this, an analysis of sensitivity of the parameters was performed and, for this study, only half of the parameters were able to cause some interference in the type of classification desired, the others were kept in their standard configuration.

```
sklearn.svm.LinearSVC(
                    penalty='l2', {"l1", "l2"} "Specifies the norm used in the penalization."
                    loss='squared_hinge', {"hinge", "squared_hinge"} "Specifies the loss function."
                    dual=True, {"True", "False"} "Select the algorithm to either solve the dual or primal optimization problem."
                    tol=0.0001, "Tolerance for stopping criteria."
                    C=1.0, "Penalty parameter C of the error term."
                    multi_class='ovr', {"ovr","crammer_singer"}"Determines the multi-class strategy if y contains more than two classes."
                    fit_intercept=True,
                    intercept_scaling=1,
                    class_weight=None,
                    verbose=0,
                    random_state=None,
                    max_iter=1000
                    )
```

Figure 5: Parameters of the LinearSVC function of the SVM classifier. The first 6 parameters listed in the figure along with a brief explanation of their changes and possible assignments in the function in question constitute the set of parameters that cause some interference in the type of result obtained.

For the analysis of the sensitivity of these parameters, 20 different scenarios were defined, -i.e., 20 combinations of the possible parameter sets and, thus, the best obtained classifier model was determined. Table 1 describes the 20 scenarios evaluated.

Table 1: LinearSVC function parameter settings used for sensitivity analysis

| Parameter Sets / Scenario | C | Dual | Loss | Multi_Class | Penalty | Tol |
|---|---|---|---|---|---|---|
| S 0 | 1 | True | squared_hinge | ovr | l2 | 0.0001 |
| S 1 | 1 | True | squared_hinge | crammer_singer | l2 | 0.0001 |
| S 2 | 1 | True | hinge | ovr | l2 | 0.0001 |
| S 3 | 1 | False | squared_hinge | ovr | l2 | 0.0001 |
| S 4 | 1 | False | squared_hinge | ovr | l1 | 0.0001 |
| S 5 | 150 | True | squared_hinge | ovr | l2 | 0.0001 |
| S 6 | 150 | True | squared_hinge | crammer_singer | l2 | 0.0001 |
| S 7 | 150 | True | hinge | ovr | l2 | 0.0001 |
| S 8 | 150 | False | squared_hinge | ovr | l2 | 0.0001 |
| S 9 | 150 | False | squared_hinge | ovr | l1 | 0.0001 |
| S 10 | 1 | True | squared_hinge | ovr | l2 | 0.01 |
| S 11 | 1 | True | squared_hinge | crammer_singer | l2 | 0.01 |
| S 12 | 1 | True | hinge | ovr | l2 | 0.01 |
| S 13 | 1 | False | squared_hinge | ovr | l2 | 0.01 |
| S 14 | 1 | False | squared_hinge | ovr | l1 | 0.01 |
| S 15 | 150 | True | squared_hinge | ovr | l2 | 0.01 |
| S 16 | 150 | True | squared_hinge | crammer_singer | l2 | 0.01 |
| S 17 | 150 | True | hinge | ovr | l2 | 0.01 |
| S 18 | 150 | False | squared_hinge | ovr | l2 | 0.01 |
| S 19 | 150 | False | squared_hinge | ovr | l1 | 0.01 |

* Underline parameters represent non-standard setting

*2.2.3. SVM Classifier Evaluation*

The evaluation of the best generalization capacity by the parameter sets was performed using the cross-validation technique so the database was partitioned into 5 samples created to test each of the calibration parameter settings. Thus, the set of parameters with the best classification average of the 5 samples tested was chosen.

*2.2.4. Generation of Training and Test Samples and Subsets*

As the purpose of the research is to identify images of motorcycle type vehicles, the 721 examples of this class were used and defined as standard quantity. Therefore, only one part was selected for the composition of each of the 5 samples generated, and a random sampling and replacement of the images were performed.

Thus, each of the 5 samples produced had the same number of examples of both vehicle classes (721 images), totaling 1,442 images per sample generated.

Each sample was subdivided into two subsets: (i) training with 1,010 vehicle images (70% of total sample images) being 505 images of each of the two classes and (ii) testing with 432 images of vehicles (30% of the total images of the sample) being 216 images of each of the two classes.

*2.2.5. Evaluation of test subset*

Figure 7 details the steps used to evaluate the 5 samples: (i) each sample was subdivided into a training subset and a test subset, (ii) for each training subset created the 20 scenarios, parameter settings of calibration, resulting in 20 classification models, each generated due to the type of scenario used, totalizing 100 models (20 classifiers trained times 5 subset of training), (iii) for each test subset the 20 classification models resulting from the previous stage producing 20 sets of predictions, each generated due to the classification model used, totaling 100 sets of predictions (20 sets of predictions times 5 test subset).

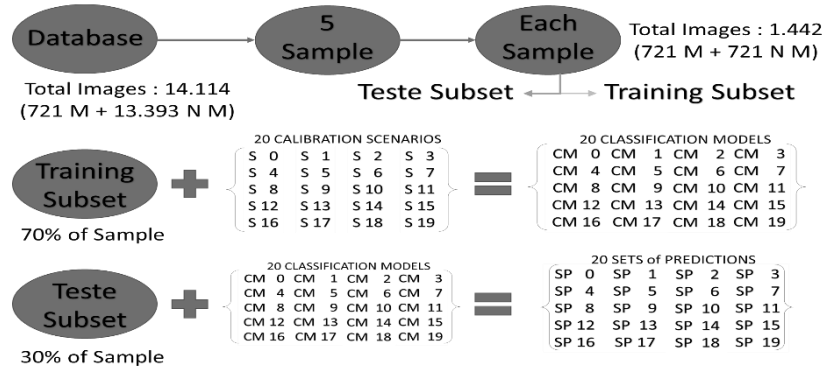

*Figure 6:* Schematic of the stages used for the created subsets.

To evaluate the performance of each of the 20 calibration parameter configurations used, each of the 100 prediction sets was generated in the contingency table, Table 2, in order to analyze: (i) the true positive rate (TPR), which represents the number of images correctly classified as motorcycles, (ii) the false positive rate (FPR), which represents the number of images mistakenly classified as motorcycles, (iii) the true negative rate (TNR), which represents the number of images correctly classified as non-motorcycles, (iv) precision, (v) accuracy, and (vi) the value of the AUC metric, which evaluates the overall prediction performance of the classifier at random (value close to 0.5) or not random (the closer to 1.0, the better overall performance) (Fawcett, 2006).

Table 2: Contingency table and metrics used to analyze the performance of the SVM classifier.

| $TP_{rate} = \dfrac{TP}{P}$  $Precision = \dfrac{TP}{Yes}$ | | TRUE CLASS | | |
|---|---|---|---|---|
| $FP_{rate} = \dfrac{FP}{N}$  $Accuracy = \dfrac{TP+TN}{P+N}$ | | MOTORCYCLE (POSITIVE) | NO MOTORCYCLE (NEGATIVE) | TOTAL |
| $TN_{rate} = \dfrac{TN}{N}$ | | | | |
| PREDICTION CLASS | MOTORCYCLE (POSITIVE) | True Positives (TP) | False Positives (FP) | Yes |
| | NO MOTORCYCLE (NEGATIVE) | False Negatives (FN) | True Negatives (TN) | No |
| | TOTAL | P | N | |

## 3.  RESULS

Through the analysis of the contingency tables of each of the 100 prediction sets, it was possible to obtain an average for each type of calibration test and to verify that only the parameters C (penalty for error) and Tol (tolerance for the criterion of stop) which showed a significant change in the predictions obtained, as shown in Table 3. The highlighted values represent the highest true positive rates (TPR) and the lowest false positive rates. (FPR)

Table 3: True positive and false positive rates of the 5 samples for each of the 20 scenarios highlighting the two main parameters (C and Tol) that demonstrated a significant difference in the prediction.

| Calibration | | TOL = $1\times10^{-4}$ | | Calibration | TOL = $1\times10^{-2}$ | |
|---|---|---|---|---|---|---|
| | | TP rate | FP rate | | TP rate | FP rate |
| C = 1 | CCP 0 | 0.916 | 0.273 | CCP 10 | 0.916 | 0.272 |
| | CCP 1 | 0.933 | 0.445 | CCP 11 | 0.933 | 0.444 |
| | CCP 2 | 0.945 | 0.538 | CCP 12 | 0.945 | 0.537 |
| | CCP 3 | 0.916 | 0.273 | CCP 13 | 0.916 | 0.273 |
| | CCP 4 | 0.927 | 0.157 | CCP 14 | 0.924 | 0.156 |
| C = 150 | CCP 5 | 0.937 | 0.098 | CCP 15 | 0.929 | 0.092 |
| | CCP 6 | 0.939 | 0.103 | CCP 16 | 0.939 | 0.102 |
| | CCP 7 | 0.930 | 0.108 | CCP 17 | 0.933 | 0.119 |
| | CCP 8 | 0.933 | 0.097 | CCP 18 | 0.934 | 0.098 |
| | CCP 9 | 0.917 | 0.085 | CCP 19 | 0.922 | 0.089 |

Accordingly to the analysis it is possible to verify that for the 2 main parameters of calibration of the function in question the variation presented in the experiment is very small which shows the great difficulty in estimating the best values of the parameters of an algorithm, corroborating the information in the literature (Arróspide & Salgado, 2014 ; S. Kim, Yu, Man, & Lee, 2015 ; Luo et al., 2018 ; Minh, Sano, & Matsumoto, 2012). According to R. Silva et al. (2013) as the values are mostly for true positive rate (TPR) above 0.9 and for false positive rate (FPR) are less than 0.1, the classifier can be considered to perform well. Therefore, the AUC metric was used to assist in choosing the best classifier calibration.

Table 4 shows the average performance of each of the test sets for both the AUC metric and true positive, false positive, true negative, as well as precision and accuracy rates. For the AUC, the best performance was 0.96, which was in accordance with FAWCETT, (2006) e PRATI;

BATISTA; MONARD (2008) for being a value higher than 0.95 allows to consider that the performance of the classifier is great and its general prediction does not have random behavior. Through the analysis of the metrics obtained, the classifier presented good general performance (accuracy and precision values above 0.91), because according to the literature (Arróspide & Salgado, 2014 ; Fawcett, 2006 ; Luo et al., 2018 ; Minh et al., 2012 ; Prati et al., 2008 ; R. Silva et al., 2013 ; R. R. V. e Silva et al., 2018), a classifier with precision values greater than 0.94, and accuracy values greater than 0.97 is an excellent performance classifier.

Table 4: Performance of the AUC metric of the 5 samples for each of the 20 calibration parameter configurations of the test sets and their usual metrics.

|  | SAMPLE 1 | SAMPLE 2 | SAMPLE 3 | SAMPLE 4 | SAMPLE 5 | AVERAGE_ AUC | TP rate | FP rate | Precision | Accuracy | TN rate |
|---|---|---|---|---|---|---|---|---|---|---|---|
| AUC_TE_M0 | 0.894 | 0.936 | 0.892 | 0.896 | 0.896 | 0.903 | 0.916 | 0.273 | 0.771 | 0.821 | 0.727 |
| AUC_TE_M1 | 0.844 | 0.898 | 0.817 | 0.848 | 0.855 | 0.852 | 0.933 | 0.445 | 0.677 | 0.744 | 0.555 |
| AUC_TE_M2 | 0.807 | 0.867 | 0.770 | 0.804 | 0.821 | 0.814 | **0.945** | 0.538 | 0.637 | 0.704 | 0.462 |
| AUC_TE_M3 | 0.894 | 0.936 | 0.892 | 0.896 | 0.896 | 0.903 | 0.916 | 0.273 | 0.771 | 0.821 | 0.727 |
| AUC_TE_M4 | 0.910 | 0.962 | 0.955 | 0.936 | 0.938 | 0.940 | 0.927 | 0.157 | 0.855 | 0.885 | 0.843 |
| AUC_TE_M5 | 0.943 | 0.973 | 0.975 | 0.963 | 0.955 | **0.962** | 0.937 | 0.098 | 0.906 | **0.919** | 0.902 |
| AUC_TE_M6 | 0.945 | 0.974 | 0.974 | 0.962 | 0.954 | 0.962 | 0.939 | 0.103 | 0.902 | 0.918 | 0.897 |
| AUC_TE_M7 | 0.942 | 0.972 | 0.970 | 0.958 | 0.951 | 0.959 | 0.930 | 0.108 | 0.896 | 0.911 | 0.892 |
| AUC_TE_M8 | 0.943 | 0.973 | 0.975 | 0.963 | 0.955 | **0.962** | 0.933 | 0.097 | 0.906 | 0.918 | 0.903 |
| AUC_TE_M9 | 0.947 | 0.970 | 0.974 | 0.966 | 0.952 | 0.962 | 0.917 | **0.085** | 0.916 | 0.916 | 0.915 |
| AUC_TE_M10 | 0.894 | 0.936 | 0.892 | 0.896 | 0.896 | 0.903 | 0.916 | 0.272 | 0.771 | 0.822 | 0.728 |
| AUC_TE_M11 | 0.844 | 0.898 | 0.817 | 0.848 | 0.855 | 0.852 | 0.933 | 0.444 | 0.678 | 0.744 | 0.556 |
| AUC_TE_M12 | 0.807 | 0.868 | 0.770 | 0.804 | 0.821 | 0.814 | **0.945** | 0.537 | 0.638 | 0.704 | 0.463 |
| AUC_TE_M13 | 0.894 | 0.936 | 0.892 | 0.896 | 0.896 | 0.903 | 0.916 | 0.273 | 0.771 | 0.821 | 0.727 |
| AUC_TE_M14 | 0.924 | 0.962 | 0.954 | 0.939 | 0.938 | 0.944 | 0.924 | 0.156 | 0.855 | 0.884 | 0.844 |
| AUC_TE_M15 | 0.943 | 0.973 | 0.975 | 0.963 | 0.955 | 0.962 | 0.929 | 0.092 | **0.911** | **0.919** | **0.908** |
| AUC_TE_M16 | 0.945 | 0.974 | 0.974 | 0.962 | 0.954 | 0.962 | 0.939 | 0.102 | 0.902 | **0.919** | 0.898 |
| AUC_TE_M17 | 0.942 | 0.972 | 0.970 | 0.958 | 0.951 | 0.959 | 0.933 | 0.119 | 0.887 | 0.907 | 0.881 |
| AUC_TE_M18 | 0.944 | 0.973 | 0.975 | 0.963 | 0.955 | **0.962** | 0.934 | 0.098 | 0.905 | 0.918 | 0.902 |
| AUC_TE_M19 | 0.949 | 0.971 | 0.975 | 0.963 | 0.953 | **0.962** | 0.922 | 0.089 | 0.913 | 0.917 | 0.911 |
| MAX | <u>0.949</u> | <u>0.974</u> | <u>0.975</u> | <u>0.966</u> | <u>0.955</u> | | | | | | |
| MIN | <u>0.807</u> | <u>0.867</u> | <u>0.770</u> | <u>0.804</u> | <u>0.821</u> | | | | | | |

## 4. CONCLUSIONS

The research in matter allowed the creation of a stock of motorcycle-type vehicles containing 721 images extracted from monitoring videos producing clear, precise and back-to-back examples of motorcycles during the urban flow of a Brazilian city.

As a major contribution, the research enabled the development and calibration of a tool that assists in the data collection of urban traffics that uses a linear separation criterion to train the classifier (LinearSVC algorithm) in conjunction with feature vectors created from histograms in gray scale by the LBP function (local_binary_pattern). The calibration of the classifier obtained by altering the parameters C = 150 and Tol = 0.01 and maintaining the other parameters in its standard configuration allowed this one to present a performance (precision and accuracy superior to 0.91 and AUC higher than 0.95) similar to the classifiers considered optimal by the literature.

Therefor the idea that it is possible to develop a motorcycle classifier of good performance using examples of images extracted from traffic monitoring videos has been proven.